\documentclass[conference]{IEEEtran}
\IEEEoverridecommandlockouts
\usepackage{cite}
\usepackage{url}
\usepackage{amsmath,amssymb,amsfonts}
\usepackage{algorithmic}
\usepackage{graphicx}
\usepackage{textcomp}
\usepackage{float}
\usepackage{xcolor}

\def\BibTeX{{\rm B\kern-.05em{\sc i\kern-.025em b}\kern-.08em
    T\kern-.1667em\lower.7ex\hbox{E}\kern-.125emX}}
\begin{document}

\title{Comprehensive Modeling Approaches for Forecasting Bitcoin Transaction Fees: A Comparative Study  \\

}

\author{\IEEEauthorblockN{1\textsuperscript{sd} Jiangqin Ma}
\IEEEauthorblockA{
\textit{Department of Data Science} \\
\textit{Blockonomics} \\
Richland, USA \\
jiangqin.ma@gmail.com}

\and
\IEEEauthorblockN{2\textsuperscript{nd} Erfan Mahmoudinia}
\IEEEauthorblockA{\textit{Department of Data Science} \\
\textit{Blockonomics} \\
 Tehran, Iran \\
mahmoudinia.erfan@gmail.com}

}

\maketitle

\begin{abstract}
Transaction fee prediction in Bitcoin's ecosystem represents a crucial challenge affecting both user costs and miner revenue optimization. This study presents a systematic evaluation of six predictive models for forecasting Bitcoin transaction fees across a 24-hour horizon (144 blocks): SARIMAX, Prophet, Time2Vec, Time2Vec with Attention, a Hybrid model combining SARIMAX with Gradient Boosting, and the Temporal Fusion Transformer (TFT). Our approach integrates comprehensive feature engineering spanning mempool metrics, network parameters, and historical fee patterns to capture the multifaceted dynamics of fee behavior.

Through rigorous 5-fold cross-validation and independent testing, our analysis reveals that traditional statistical approaches outperform more complex deep learning architectures. The SARIMAX model achieves superior accuracy on the independent test set, while Prophet demonstrates strong performance during cross-validation. Notably, sophisticated deep learning models like Time2Vec and TFT show comparatively lower predictive power despite their architectural complexity. This performance disparity likely stems from the relatively constrained training dataset of 91 days, suggesting that deep learning models may achieve enhanced results with extended historical data.

These findings offer significant practical implications for cryptocurrency stakeholders, providing empirically-validated guidance for fee-sensitive decision making while illuminating critical considerations in model selection based on data constraints. The study establishes a foundation for advanced fee prediction while highlighting the current advantages of traditional statistical methods in this domain.
\end{abstract}


\section{Introduction}
In the dynamic and volatile cryptocurrency ecosystem, the accurate prediction of transaction fees represents a critical challenge for both individual users and institutional stakeholders \cite{easley2019bitcoin}\cite{houy2014bitcoin}. Bitcoin transaction fees exhibit significant fluctuations influenced by multiple factors, including mempool size, transaction volume, and block time \cite{moser2015trends} \cite{kasahara2019effect}. Effective fee rate prediction enables users to optimize transaction timing, thereby enhancing both cost-effectiveness and operational efficiency \cite{möser2020empirical}.

To address this challenge, we develop and evaluate predictive models for forecasting Bitcoin's median fee rate over a 24-hour horizon (corresponding to 144 blocks). Our analysis leverages a comprehensive dataset spanning 91 days (September 16, 2024 to December 15, 2024), comprising 11,809 records with 23 distinct features organized into four categories: Transaction and Mempool Metrics, Block Information, Network Parameters, and Temporal Features \cite{median2023bitcoin}. This rich feature set captures the multifaceted nature of Bitcoin network dynamics, from fundamental transaction metrics to broader system-level indicators \cite{blockchain2023metrics}.

The production pipeline initiates with data acquisition from the Bitcoin Core server \cite{bitcoincore}, which provides essential network metrics including block height, mempool size, difficulty, and hash rate \cite{mempool2023bitcoin}\cite{blockchair}. We augment this data by integrating real-time Bitcoin price information from the CoinGecko API \cite{coingecko}. Our preprocessing strategy employs forward and backward filling for missing values and feature-specific percentile-based clipping for outliers, maintaining data integrity while preserving temporal sequence.

For model evaluation, we implement a rigorous framework using 11,809 records for training and validation through 5-fold cross-validation, while reserving 144 records (24 hours) as a consistent test set \cite{bergstra2012random}. We evaluate six distinct models: Prophet \cite{taylor2018forecasting}, Time2Vec \cite{kazemi2019time2vec}, Time2Vec with Attention \cite{vaswani2017attention}, SARIMAX \cite{box2015time}, a Hybrid model combining SARIMAX with Gradient Boosting \cite{friedman2001greedy}, and the Temporal Fusion Transformer \cite{lim2021temporal}. This comprehensive approach enables thorough model validation while ensuring realistic performance assessment for practical fee prediction applications.

The dataset and code used in this study are publicly available in a GitHub repository \cite{githubrepo}, ensuring reproducibility and transparency.

\section{Related Work}
Research in cryptocurrency fee prediction has evolved significantly, spanning multiple methodological approaches. We organize the relevant literature into three main categories: economic analysis, time series approaches, and advanced statistical methods.

\subsection{Economic Analysis of Transaction Fees}
Early work in cryptocurrency fee analysis focused on economic fundamentals. Easley et al.~\cite{easley2019bitcoin} provided foundational insights into Bitcoin fee markets, establishing the relationship between transaction fees and mining economics. Their research demonstrated how fee dynamics emerge from the interaction between users and miners, though their analysis focused on equilibrium patterns rather than prediction. Houy~\cite{houy2014bitcoin} developed one of the first analytical frameworks for Bitcoin fee economics, proposing models based on game theory and market equilibrium.

Building on these economic foundations, Moser and Böhme~\cite{moser2015trends} conducted longitudinal analysis of Bitcoin transaction fees, identifying key patterns in fee rate evolution. Kasahara and Kawahara~\cite{kasahara2019effect} extended this work by examining the effect of fees on transaction confirmation times, introducing temporal considerations that inform current predictive approaches.

\subsection{Time Series Modeling}
The application of time series analysis to cryptocurrency data marked a significant advancement in fee prediction methodology. Box et al.~\cite{box2015time} established the statistical foundations for analyzing temporal patterns in financial markets, providing techniques that would later prove crucial for cryptocurrency analysis. Building on these methods, Möser and Böhme~\cite{möser2020empirical} expanded the understanding of Bitcoin transaction dynamics through empirical analysis of fee structures and network characteristics, demonstrating key relationships between network metrics and transaction costs.

Recent studies have extended time series approaches to cryptocurrency forecasting. For example, 
Estimating and forecasting bitcoin daily prices using ARIMA-GARCH models.
Phung Duy et al.~\cite{kamalov2021forecasting} explored hybrid ARIMA-GARCH models for Bitcoin price prediction, which offer insights for fee forecasting due to their shared temporal characteristics. Similarly, Fischer and Krauss~\cite{fischer2018deep} utilized LSTM networks for financial forecasting, demonstrating their ability to capture long-term dependencies and volatility in sequential data.

\subsection{Advanced Statistical Methods}
Recent developments in machine learning and explainable AI have enabled more sophisticated approaches to cryptocurrency analysis. Vaswani et al.~\cite{vaswani2017attention} introduced transformer architectures that revolutionized temporal data processing, while Kazemi et al.~\cite{kazemi2019time2vec} developed temporal embeddings specifically designed for time series analysis. These advances in temporal modeling were complemented by Friedman's~\cite{friedman2001greedy} gradient boosting framework, which has proven particularly effective for cryptocurrency applications.

Transformer-based models have gained significant attention for financial forecasting. Lim et al.~\cite{lim2021temporal} introduced the Temporal Fusion Transformer (TFT), which remains a cornerstone for interpretable multi-horizon time series forecasting. More recently, Muhammad et al.~\cite{muhammad2023transformer} demonstrated how attention-based transformers can be applied to stock market prediction, emphasizing feature importance and interpretability. These approaches align with advancements in explainable AI, as discussed by Arrieta et al.~\cite{arrieta2020explainable}, ensuring transparency in model predictions.

Hybrid approaches have also gained traction. Vargas et al.~\cite{vargas2018hybrid} proposed a hybrid approach combining convolutional neural networks (CNNs) and LSTMs for cryptocurrency price prediction, showing how feature extraction and sequential modeling can be integrated effectively. Similarly, Chong et al.~\cite{chong2017deep} utilized hybrid deep learning frameworks for financial time series forecasting, capturing both short-term trends and long-term dependencies.

Our research builds upon these foundations while addressing several critical gaps in the literature. Most existing studies focus on equilibrium analysis or price prediction rather than dynamic fee forecasting. Additionally, the interaction between network metrics, market conditions, and fee rates remains inadequately modeled for longer time horizons. Our work extends both the theoretical understanding and practical applications of fee prediction while incorporating comprehensive feature engineering, hybrid modeling, and transformer-based approaches.

\section{Dataset Overview} 
Our analysis leverages a comprehensive dataset spanning September 16, 2024, to December 15, 2024, encompassing 11,809 records over approximately 91 days. While Bitcoin's protocol targets an average block time of 10 minutes, actual intervals vary due to network difficulty adjustments and inherent mining randomness. Each record captures 23 distinct features characterizing Bitcoin network dynamics.

\subsection{Data Preprocessing} 
To ensure data quality and preserve the temporal sequence, we implemented a robust preprocessing strategy. Duplicate values were removed to maintain consistency. Missing values were addressed using forward and backward filling methods, ensuring completeness without disrupting temporal continuity. Outliers were managed using feature-specific percentile-based clipping, allowing for mitigation of extreme values while retaining the full sample size.

For model evaluation, we allocated 11,665 records (11,809 - 144) for training and validation, implementing 5-fold cross-validation across six models: Prophet, Time2Vec, Time2Vec with Attention, SARIMAX, a SARIMAX-Gradient Boosting hybrid, and the Temporal Fusion Transformer (TFT). The remaining 144 records were reserved for the test set to provide a realistic assessment of 24-hour fee prediction performance.

\subsection{Feature Categories}
Our features are organized into four key categories:

\subsubsection{Transaction and Mempool Metrics}
These features capture network transactional dynamics through indicators such as transaction count, mempool size, and a comprehensive set of fee statistics. These include minimum, maximum, average, and median values, as well as the 10th and 90th percentiles and standard deviations to represent fee distribution characteristics.

\subsubsection{Block Information}
This category includes blockchain-specific parameters such as block height, block weight, block interval, and block version. These features provide structural and temporal insights into the blockchain.

\subsubsection{Network Parameters and Feature Engineering}
This category encompasses system-level indicators such as mining difficulty, hash rate, and Bitcoin's USD price. Additional features derived from mempool fee histograms include metrics like low, medium, and high fee ratios, as well as a fee diversity measure that captures the dispersion of fees across the network.

\subsubsection{Temporal Features}
Temporal features include timestamp data and processed mempool fee histograms. The target variable, \texttt{block\_median\_fee\_rate}, represents the median fee rate per block and serves as the primary prediction objective.

\subsection{Feature Correlation Analysis}
\begin{figure}[htbp]
\centerline{\includegraphics[width=0.5\textwidth]{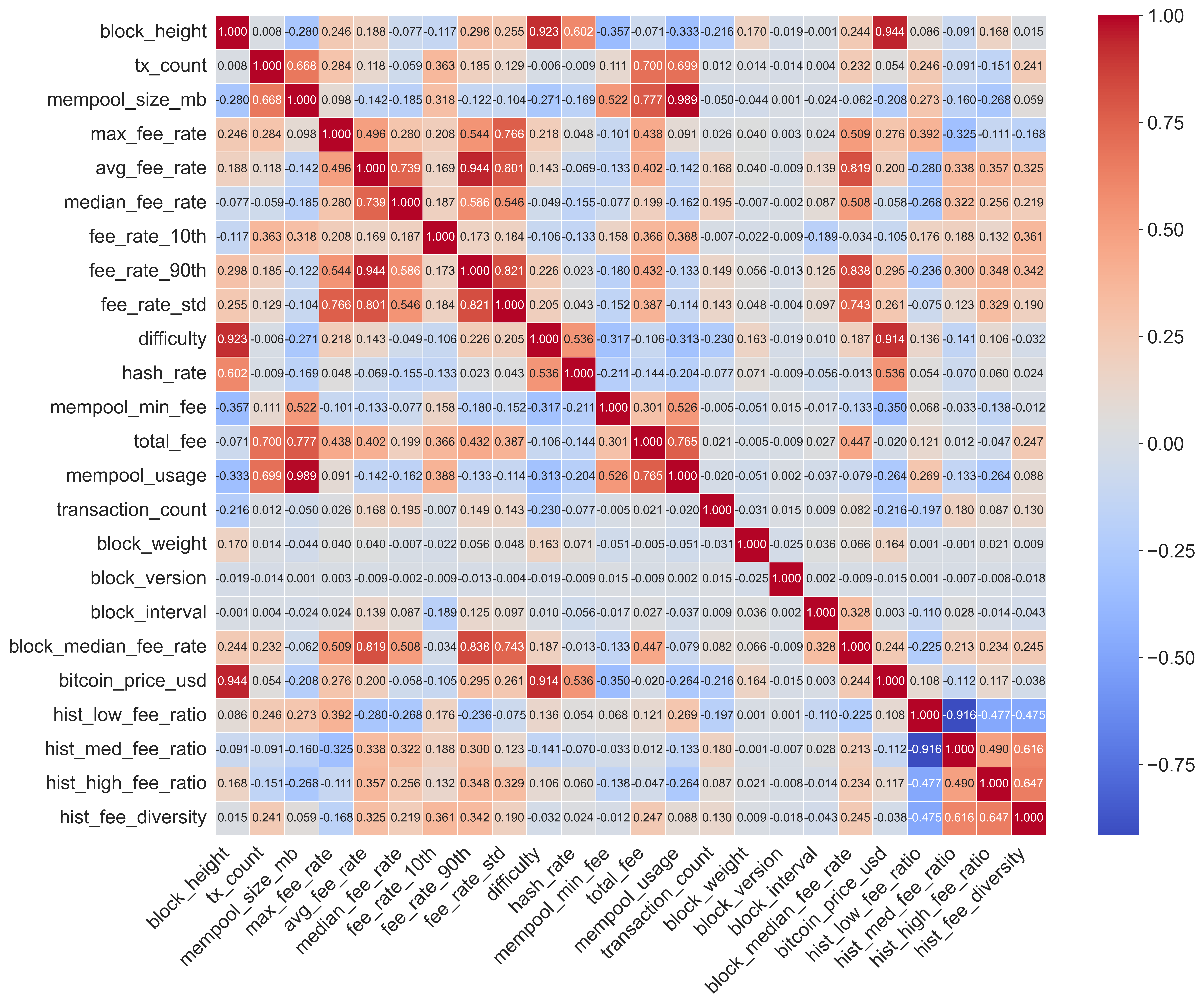}}
\caption{Correlation Heatmap of Features.}
\label{fig}
\end{figure}

The correlation heatmap in Figure~\ref{fig} highlights significant relationships among the 23 features. Several key patterns emerge:

\begin{itemize}
\item \textbf{Fee Rate Correlations:} The mempool's fee-related metrics show strong positive correlations among themselves. The average fee rate (\texttt{avg\_fee\_rate}) strongly correlates with both the 90th percentile fee rate (\texttt{fee\_rate\_90th}, $r = 0.944$) and fee rate standard deviation (\texttt{fee\_rate\_std}, $r = 0.801$). Similarly, the block median fee rate (\texttt{block\_median\_fee\_rate}) demonstrates notable correlations with both Bitcoin price (\texttt{bitcoin\_price\_usd}, $r = 0.244$) and average fee rate (\texttt{avg\_fee\_rate}, $r = 0.819$).

\item \textbf{Mempool Size Relationships:} The mempool size in megabytes (\texttt{mempool\_size\_mb}) shows inverse relationships with fee metrics, displaying negative correlations with both average fee rate (\texttt{avg\_fee\_rate}, $r = -0.142$) and median fee rate (\texttt{median\_fee\_rate}, $r = -0.185$). This inverse relationship indicates that larger mempool sizes tend to coincide with lower fee rates.

\item \textbf{Fee Distribution Dynamics:} The analysis of histogram-based fee metrics reveals significant patterns in fee distribution. Most notably, there's a strong negative correlation between low-fee and medium-fee transaction ratios (\texttt{hist\_low\_fee\_ratio} and \texttt{hist\_med\_fee\_ratio}, $r = -0.916$), indicating that an increase in low-fee transactions corresponds to a decrease in medium-fee transactions.

\item \textbf{Independent Features:} Certain blockchain characteristics, including block version (\texttt{block\_version}) and hash rate (\texttt{hash\_rate}), demonstrate low correlations with other variables ($|r| < 0.3$), suggesting they provide independent predictive signals.
\end{itemize}

\subsection{Temporal Dynamics Analysis}
Figure~\ref{fig:bitcoin_metrics_revised} illustrates the temporal dynamics of key network metrics over a 48-hour observation period (2024-10-01 to 2024-10-02), revealing several critical patterns in Bitcoin network behavior.
Transaction activity demonstrates substantial variability, with counts fluctuating between 1,000 and 8,000 transactions per block, accompanied by corresponding block intervals ranging from near-instantaneous to 1,400 seconds. This variability reflects the network's dynamic response to changing transaction loads.

The mempool metrics reveal patterns of network congestion, with the average fee rate maintaining stability around 1.6 units but experiencing spikes up to 2.8 units during high-activity periods. The mempool size exhibits a characteristic V-shaped pattern, decreasing from 48 MB to 38 MB before recovering to 50 MB, indicating cycles of transaction processing and accumulation.

Network performance indicators show contrasting trends: the hash rate demonstrates steady growth from 6.0 to 7.5 units, while Bitcoin price declines from 63,500 to 60,500 USD. The target variable, block median fee rate, fluctuates significantly between 2 and 10 units, with variations closely tied to transaction volume and block interval patterns.

These observations highlight the complex interplay between network metrics and suggest the need for predictive models that can effectively capture both immediate correlations and lagged effects across multiple variables.

\section{Methodology}
While extensive research has been conducted on Bitcoin price prediction \cite{mcnamara2019predicting}, transaction fee forecasting remains relatively underexplored in the academic literature. This research gap motivated the application of advanced time-series modeling approaches to address the unique challenges of fee prediction in the volatile cryptocurrency ecosystem. Six distinct models representing different paradigms in time-series forecasting were evaluated in this study:

\begin{itemize}
    \item \textbf{Prophet} \cite{taylor2018forecasting}, designed for business time series, is particularly adept at handling multiple seasonalities and holidays.
    \item \textbf{Time2Vec} \cite{kazemi2019time2vec} and its attention-enhanced variant \cite{vaswani2017attention}, which capture complex temporal dependencies inherent in fee patterns.
    \item \textbf{SARIMAX} (Seasonal AutoRegressive Integrated Moving Average with eXogenous variables) \cite{box2015time}, capable of modeling both seasonal patterns and external features.
    \item A \textbf{Hybrid model} that combines SARIMAX with Gradient Boosting, leveraging the strengths of both statistical and machine learning approaches \cite{friedman2001greedy}.
    \item \textbf{Temporal Fusion Transformer (TFT)}, an advanced deep learning architecture integrating variable selection with multi-horizon forecasting capabilities \cite{lim2021temporal}.
\end{itemize}

\subsection{Model Architectures}

\subsubsection{SARIMAX}
The SARIMAX model incorporates both seasonal patterns and external explanatory variables, making it suitable for capturing the cyclical behavior of Bitcoin fee rates. Configured with parameters \( (p, d, q) \times (P, D, Q)_s \), where \( p = 2 \), \( d = 1 \), \( q = 2 \), \( P = 1 \), \( D = 1 \), and \( Q = 1 \), the model accounts for daily seasonality (\( s = 144 \), corresponding to 144 Bitcoin blocks). This configuration effectively models periodic trends and autoregressive dependencies in the fee rate dynamics \cite{box2015time}.

\subsubsection{Time2Vec}
Time2Vec transforms temporal data into higher-dimensional representations, enabling the capture of multiple periodic patterns across different frequencies. The architecture includes a 64-dimensional embedding layer followed by three dense layers with 128, 64, and 32 units, respectively \cite{kazemi2019time2vec}. The attention-enhanced variant integrates a multi-head self-attention mechanism \cite{vaswani2017attention}, using 8 heads to capture long-range dependencies and interactions in fee rates.

\subsubsection{Prophet}
Prophet, a model developed by Facebook, is designed to handle time-series data with strong seasonal components. It is particularly effective for modeling non-linear trends and seasonalities. In this study, Prophet was configured to capture daily seasonality and employed automatic changepoint detection to account for abrupt changes in the Bitcoin fee rate trends \cite{taylor2018forecasting}.

\subsubsection{Temporal Fusion Transformer (TFT)}
The Temporal Fusion Transformer (TFT) architecture combines variable selection networks, gated residual networks, and multi-head attention mechanisms to generate interpretable multi-horizon forecasts. For this study, the TFT model utilized four attention heads, a hidden dimension of 128, and a dropout rate of 0.1 \cite{lim2021temporal}. This configuration enables effective modeling of both short-term and long-term temporal dependencies in Bitcoin fee rates.

\subsection{Hybrid Model Architecture}
The Hybrid model combines the strengths of SARIMAX and Gradient Boosting to address the non-linear and seasonal complexities of Bitcoin fee rate prediction. The architecture operates in two stages (see Fig.~\ref{fig:algorithm1} for the algorithmic details):

\textbf{Stage 1: SARIMAX Component.} \\
SARIMAX captures linear relationships and seasonal patterns while incorporating exogenous variables such as mempool size and transaction count. The model produces base predictions, \(\hat{y}_s\), and residuals, \(r_s = y - \hat{y}_s\), which quantify errors in prediction.

\textbf{Stage 2: Gradient Boosting Component.} \\
The Gradient Boosting component, implemented using LightGBM, processes an enhanced feature set that includes SARIMAX predictions, residuals, rolling statistics, and lagged variables of key features. Early stopping is employed to prevent overfitting during training.

\textbf{Integration Mechanism.} \\
The final prediction combines the outputs of SARIMAX and Gradient Boosting through a dynamically weighted average:
\begin{equation}
\text{Final Prediction:} \quad \hat{y} = \alpha \cdot \hat{y}_s + (1 - \alpha) \cdot \hat{y}_g
\label{eq:hybrid_prediction}
\end{equation}
where the weight \(\alpha\) is adjusted dynamically using an exponential moving average (EMA) of recent prediction errors:
\begin{equation}
\alpha = \frac{1}{1 + \exp(\textsc{EMA}(e_s) - \textsc{EMA}(e_g))}
\end{equation}

The hybrid model's dynamic weighting mechanism allows it to adapt to changing network conditions, leveraging SARIMAX’s strength in capturing trends and Gradient Boosting’s ability to model non-linear interactions. This approach improves robustness and accuracy across varying market conditions. For a step-by-step implementation, refer to Fig.~\ref{fig:algorithm1}.

\begin{figure}[!t]
\centering
\fbox{%
\begin{minipage}{0.95\linewidth}
\small % Set font size to small (9pt) for algorithms
\textbf{Hybrid Fee Rate Prediction Model Algorithm} \\[2pt]
\hrule
\vspace{4pt}

\textbf{Require:} Input Parameters: \\
\(\mathbf{X_{t-n:t}}\): Historical feature matrix from \(t - n\) to \(t\), \\
\(\mathbf{y_{t-n:t}}\): Historical fee rates, \\
\(\mathbf{w}\): Rolling window size, \\
\(\mathbf{h}\): Prediction horizon (144 blocks). \\[4pt]

\textbf{Ensure:} Output: \\
\(\hat{y}_{t+1:t+h}\): Predicted fee rates. \\[6pt]

\textbf{SARIMAX Component:}
\begin{enumerate}
    \item Initialize SARIMAX \((p, d, q) \times (P, D, Q)_s\): \\
    \(p = 2, d = 1, q = 2; P = 1, D = 1, Q = 1, s = 144\).
    \item Fit SARIMAX: \(\hat{y}_s \gets \textsc{SARIMAX.fit}(X, y)\).
    \item Compute residuals: \(r_s \gets y - \hat{y}_s\).
\end{enumerate}

\textbf{Gradient Boosting Component:}
\begin{enumerate}
    \item Construct enhanced features: \\
    \(X' \gets \textsc{ConcatenateFeatures}(X, \hat{y}_s, r_s)\).
    \item Add rolling stats/lagged features: \\
    \(X' \gets \textsc{AddRollingStatistics}(X', w)\). \\
    \(X' \gets \textsc{AddLaggedFeatures}(X')\).
    \item Configure GBM: \texttt{trees}=1000, \texttt{depth}=8, \texttt{lr}=0.01. \\
    Fit model: \(\hat{y}_g \gets \textsc{GBM.fit}(X', y)\).
\end{enumerate}

\textbf{Weight Update:}
\begin{enumerate}
    \item Compute dynamic weight: \\
    \(\alpha \gets \frac{1}{1 + \exp(\textsc{EMA}(e_s) - \textsc{EMA}(e_g))}\).
\end{enumerate}

\textbf{Prediction:}
\begin{enumerate}
    \item Combine predictions: \(\hat{y} \gets \alpha \cdot \hat{y}_s + (1 - \alpha) \cdot \hat{y}_g\).
    \item Output predicted fee rates: \(\hat{y}\).
\end{enumerate}
\vspace{4pt}
\hrule
\end{minipage}}
\caption{Hybrid Fee Rate Prediction Model Algorithm.}
\label{fig:algorithm1}
\end{figure}

\subsection{Cross-Validation Framework}
We implement a rigorous 5-fold cross-validation framework to evaluate model performance. The validation strategy employs an expanding window approach, where each successive fold incorporates additional historical data while maintaining a consistent testing horizon of 144 blocks (approximately one day). The process begins with an initial training period of 9,665 data points, and each subsequent fold expands the training window by 144 points, with the corresponding test set shifting forward by the same interval. This progressive expansion ensures that the models leverage all available historical information while preserving temporal consistency in the evaluation process.

Model performance is assessed using three metrics: Mean Absolute Error (MAE), Root Mean Square Error (RMSE), and Theil's U statistic. While MAE and RMSE provide insights into absolute and squared prediction errors, Theil's U statistic offers a scale-independent evaluation by comparing the model's accuracy against a naïve benchmark, such as a random walk or a simple historical average \cite{theil1966applied}. Theil's U is defined as:
\begin{equation}
\text{Theil's U} = \frac{\sqrt{\frac{1}{n}\sum_{t=1}^n (y_t - \hat{y}_t)^2}}{\sqrt{\frac{1}{n}\sum_{t=1}^n (y_t - y_{t-1})^2}}
\end{equation}
where \( y_t \) represents the actual value, \( \hat{y}_t \) is the predicted value, and \( n \) is the number of observations. A Theil's U value of 0 indicates perfect prediction accuracy, while a value of 1 suggests the model performs no better than the benchmark \cite{theil1966applied}.

The 5-fold cross-validation procedure spans five distinct 24-hour test periods within the dataset. Each test set provides a unique evaluation of model performance under varying market conditions and temporal patterns, ensuring that the results are robust and generalizable.

This study builds on insights from related cryptocurrency price prediction research \cite{theil1966applied}\cite{mcnamara2019predicting}, extending these methodologies to address the specific challenges of Bitcoin transaction fee forecasting. By adapting established predictive techniques to the unique characteristics of fee dynamics, including network-level metrics and market behavior patterns, this research contributes to a critical yet understudied domain within cryptocurrency analytics.

\section{Results}

\subsection{Cross-Validation Performance}

Table \ref{tab:model_performance} summarizes the average cross-validation performance of the six models. The SARIMAX model consistently demonstrated the best overall performance, achieving the lowest Mean Absolute Error (MAE) of 2.4249 and Root Mean Squared Error (RMSE) of 3.1584. This highlights its ability to balance precision and robustness across diverse market conditions, making it the most reliable model in this study.

Prophet followed closely with an MAE of 2.6924 and RMSE of 3.1890. While it excelled at capturing overall trends, it occasionally smoothed over sharp fluctuations, as shown in the cross-validation plots (Figure \ref{fig:model_comparison_appendix}). Despite these limitations, Prophet remains suitable for trend analysis and medium-term predictions.

The Hybrid model, combining SARIMAX and Gradient Boosting, exhibited moderate accuracy (MAE = 3.6012, RMSE = 4.5169). Although it effectively captured general fee rate dynamics, its predictions showed higher variance compared to SARIMAX and Prophet. This indicates sensitivity to volatile data and the potential for overfitting.

The Temporal Fusion Transformer (TFT) produced promising results in certain folds but showed inconsistent overall performance (MAE = 4.2583, RMSE = 4.8404). Its reliance on multi-head attention mechanisms and complex architectures may require larger datasets for optimal generalization.

Conversely, the Time2Vec-based models, including Time2Vec + Attention, struggled significantly in comparison. Time2Vec (MAE = 3.0138, RMSE = 3.6103) and Time2Vec + Attention (MAE = 4.0107, RMSE = 4.7562) failed to adapt effectively to the high volatility and abrupt changes in fee rates. Their overly smoothed predictions limit their applicability in dynamic environments such as Bitcoin fee rate forecasting.

In summary, SARIMAX and Prophet are the most reliable models for cross-validation, with SARIMAX demonstrating superior precision in capturing both long-term trends and short-term variability. While the Hybrid and TFT models show potential for improvement, the Time2Vec-based approaches require significant refinement to address the volatility of Bitcoin fee rates.

\begin{table}[H]
\centering
\caption{Model Performance Comparison on Average Cross-Validation Across 5 Folds}
\begin{tabular}{|l|c|c|c|}
\hline
\textbf{Model Name}          & \textbf{MAE}  & \textbf{RMSE} & \textbf{Theil's U} \\ \hline
SARIMAX                     & 2.4249        & 3.1584        & 0.5645             \\ \hline
Prophet                     & 2.6924        & 3.1890        & 0.5732             \\ \hline
Time2Vec                    & 3.0138        & 3.6103        & 0.6107             \\ \hline
Hybrid                      & 3.6012        & 4.5169        & 0.6519             \\ \hline
TFT                         & 4.2583        & 4.8404        & 0.6528             \\ \hline
Time2Vec + Attention        & 4.0107        & 4.7562        & 0.6544             \\ \hline
\end{tabular}
\label{tab:model_performance}
\end{table}

\subsection{Test Dataset Results}

Table \ref{tab:model_performance_test} presents the performance metrics of all models on the independent test dataset, which includes 144 Bitcoin blocks. Consistent with the cross-validation results, the SARIMAX model outperformed others, achieving the lowest MAE (1.2462) and RMSE (1.5859). Figure \ref{fig:model_comparison_test_appendix} further illustrates its predictive accuracy, with SARIMAX closely following the actual fee rates and exhibiting minimal deviations, even during volatile periods.

The Hybrid model, which combines SARIMAX and Gradient Boosting, followed closely with an MAE of 1.3492 and RMSE of 1.6508. While it demonstrated improved predictive accuracy by leveraging SARIMAX’s statistical robustness and Gradient Boosting’s ability to capture non-linear patterns, the Hybrid model exhibited slightly larger deviations compared to SARIMAX, particularly during abrupt market changes.

Prophet achieved an MAE of 1.5216 and RMSE of 1.9140, consistently capturing broader trends but struggling with sharp fluctuations. As seen in Figure \ref{fig:model_comparison_test_appendix}, its predictions are smoother than the actual fee rates, making Prophet more suitable for medium to long-term trend forecasting rather than precise short-term predictions.

The Temporal Fusion Transformer (TFT) model performed comparably to Prophet, with an MAE of 1.5111 and RMSE of 1.9171. However, it exhibited a tendency to oversmooth predictions during periods of high volatility, mirroring its performance during cross-validation. This highlights its challenges in adapting to rapid fee rate changes despite its sophisticated architecture.

Time2Vec and Time2Vec + Attention models continued to underperform on the test dataset. Time2Vec + Attention achieved an MAE of 1.5793 and RMSE of 1.9895 but struggled during periods of rapid change, while Time2Vec exhibited the weakest overall performance (MAE = 2.4425, RMSE = 2.7915). These results underscore the need for further architectural refinements to enhance their adaptability to volatile scenarios.

In conclusion, SARIMAX and the Hybrid model emerged as the most reliable for short-term fee prediction, offering accurate and robust predictions in a highly volatile context. Prophet and TFT demonstrated effectiveness for broader trend analysis, while Time2Vec-based models require significant refinement to achieve competitive performance in this domain.

\begin{table}[H]
\centering
\caption{Model Performance Comparison on Test Dataset (144 Blocks)}
\begin{tabular}{|l|c|c|c|}
\hline
\textbf{Model Name}          & \textbf{MAE}  & \textbf{RMSE} & \textbf{Theil's U} \\ \hline
SARIMAX                     & 1.2462        & 1.5859        & 0.4846             \\ \hline
Hybrid                      & 1.3492        & 1.6508        & 0.4927             \\ \hline
Prophet                     & 1.5216        & 1.9140        & 0.5476             \\ \hline
TFT                         & 1.5111        & 1.9171        & 0.5459             \\ \hline
Time2Vec + Attention        & 1.5793        & 1.9895        & 0.5593             \\ \hline
Time2Vec                    & 2.4425        & 2.7915        & 0.6042             \\ \hline
\end{tabular}
\label{tab:model_performance_test}
\end{table}

\section{Discussion}

\subsection{Model Complexity and Performance Relationship}
Our comprehensive analysis of six distinct predictive models reveals fundamental insights about the relationship between model complexity and prediction accuracy in cryptocurrency markets. The superior performance of traditional statistical models over more complex deep learning architectures illuminates a crucial principle: in the domain of Bitcoin fee prediction, simpler models with strong theoretical foundations can outperform sophisticated neural architectures.

The SARIMAX model's success can be attributed to several key theoretical factors. First, transaction fees demonstrate clear seasonal patterns aligned with network activity cycles, which SARIMAX explicitly models through its seasonal components. These patterns manifest at multiple frequencies: daily peaks and troughs in network congestion, weekly trading patterns, and broader market cycles. SARIMAX's ability to decompose these various seasonal elements while maintaining model interpretability proves particularly effective.

Furthermore, the relationship between key predictive features (such as mempool size and transaction volume) and fee rates appears to follow relatively stable statistical patterns over our observation period. The linear relationships captured by SARIMAX, combined with its ability to account for autoregressive and moving average components, adequately model these dynamics without requiring the additional complexity of deep neural architectures.

\subsection{Deep Learning Performance Analysis}
The underperformance of more complex models like Time2Vec and TFT can be explained through the lens of the bias-variance tradeoff. These sophisticated architectures introduce numerous parameters and non-linear relationships that, while theoretically capable of capturing more complex patterns, require substantially more data to learn stable representations. With our 91-day dataset, these models likely suffer from high variance, struggling to differentiate genuine patterns from noise in the fee rate dynamics.

This observation aligns with recent findings in financial time series prediction, where simpler models often demonstrate robust performance due to their ability to capture fundamental market mechanics without overfitting to temporary fluctuations. In cryptocurrency markets specifically, the rapid evolution of market dynamics may favor models that can adapt quickly to changing conditions over those that attempt to learn complex long-term patterns.

The Prophet model's strong performance in cross-validation further supports this hypothesis, as it combines flexible trend modeling with explicit seasonal components while maintaining relative simplicity compared to deep learning approaches. This balance between model expressiveness and complexity appears particularly well-suited to the cryptocurrency fee prediction task.

\subsection{Hybrid Model Performance and Implications}
The Hybrid model's success demonstrates how selective integration of statistical and machine learning approaches can enhance predictive power while maintaining model stability. By combining SARIMAX's trend analysis capabilities with Gradient Boosting's ability to capture non-linear relationships, the Hybrid model achieves a balance between complexity and interpretability. This suggests that future research in cryptocurrency fee prediction might benefit from focusing on hybrid approaches that maintain the interpretability and stability of traditional statistical models while selectively incorporating non-linear components only where they demonstrably improve prediction accuracy.

\subsection{Smoothing Effects and Prediction Dynamics}
A notable characteristic across all models is the presence of smoothing effects, particularly pronounced in longer-term predictions spanning 144 blocks. This smoothing phenomenon, while reducing short-term precision, represents a pragmatic compromise given Bitcoin's inherent fee rate volatility. The forecasting challenge is further complicated by cascading uncertainties from multiple exogenous variables, including mempool size, Bitcoin price fluctuations, and transaction volume patterns, each requiring their own predictive models.

The observed smoothing effect reflects an optimal balance between generalization and precision. While this approach may not capture sudden fee rate variations, it enhances the reliability of long-term strategic planning. The smoothed predictions provide valuable guidance for transaction timing decisions, offering more practical utility in volatile market conditions than precise block-by-block forecasts.

\subsection{Methodological Challenges and Limitations}
The reliance on smoothed predictions highlights fundamental challenges in forecasting volatile exogenous features. Although this approach reduces granular prediction precision, it maintains significant utility for applications prioritizing trend analysis over exact fee predictions. The models prove particularly valuable for strategic planning scenarios, such as optimizing transaction fees across multiple blocks.

The relatively small training dataset (91 days) posed particular challenges for complex architectures like TFT, which typically require extensive training data to achieve optimal performance. Additionally, the inherent volatility of Bitcoin fee rates necessitates a delicate balance between model responsiveness and prediction stability.

\subsection{Advancement Beyond Current Solutions}
Our research represents a significant advancement in Bitcoin transaction fee prediction capabilities. The key innovation lies in our models' ability to provide reliable fee predictions across a 144-block horizon (approximately 24 hours), a substantial extension beyond existing approaches. While current industry solutions focus primarily on immediate or near-term fee estimation, our work addresses the critical need for longer-term strategic fee planning.

The extended prediction window enables several practical advantages. First, it allows organizations to optimize transaction scheduling across entire operational cycles, potentially leading to substantial fee savings through improved timing decisions. Second, it provides a more comprehensive view of fee dynamics, enabling users to identify optimal transaction windows well in advance. This capability is particularly valuable for enterprises managing multiple transactions or requiring advance planning for time-sensitive operations.

Our models achieve this extended forecasting horizon while maintaining reliable accuracy through sophisticated feature engineering and advanced modeling techniques. By incorporating a rich set of network metrics, mempool characteristics, and market indicators, we capture complex fee dynamics that influence rates over longer time periods. This comprehensive approach represents a meaningful step forward in Bitcoin fee prediction technology, bridging the gap between immediate fee estimation and strategic transaction planning.

\subsection{Future Research Directions}
Future research should focus on several promising directions for enhancing predictive capabilities. First, investigating the optimal balance between model complexity and prediction accuracy through systematic ablation studies could provide valuable insights for model design. Second, exploring advanced regularization techniques for deep learning models, specifically tailored to cryptocurrency time series, could improve their performance on limited datasets.

Additional areas for investigation include:
\begin{itemize}
    \item Development of adaptive hybrid architectures that can dynamically adjust their complexity based on market conditions
    \item Integration of network-specific features such as mining difficulty adjustments and block size dynamics
    \item Exploration of transfer learning approaches to leverage knowledge from related cryptocurrency markets
    \item Investigation of interpretable deep learning architectures that maintain transparency while capturing complex market dynamics
\end{itemize}

\subsection{Practical Applications and Impact}
Our findings have significant implications for Bitcoin transaction management and fee optimization strategies. The models' smoothed predictions, while trading some precision for reliability, provide valuable guidance for decision-making in volatile market conditions. The extended prediction horizon enables more sophisticated transaction scheduling and fee optimization strategies than previously possible.

This research establishes a foundation for advanced cryptocurrency analytics while highlighting the complementary nature of short-term and long-term predictions. While SARIMAX and Hybrid models currently demonstrate superior performance, continued refinement of advanced architectures like Time2Vec and TFT promises further improvements in prediction accuracy and adaptability, advancing the broader field of cryptocurrency analytics.

\appendix
\section{Model Comparison Figures}

\begin{figure*}[t]
\centering
\includegraphics[width=0.9\textwidth]{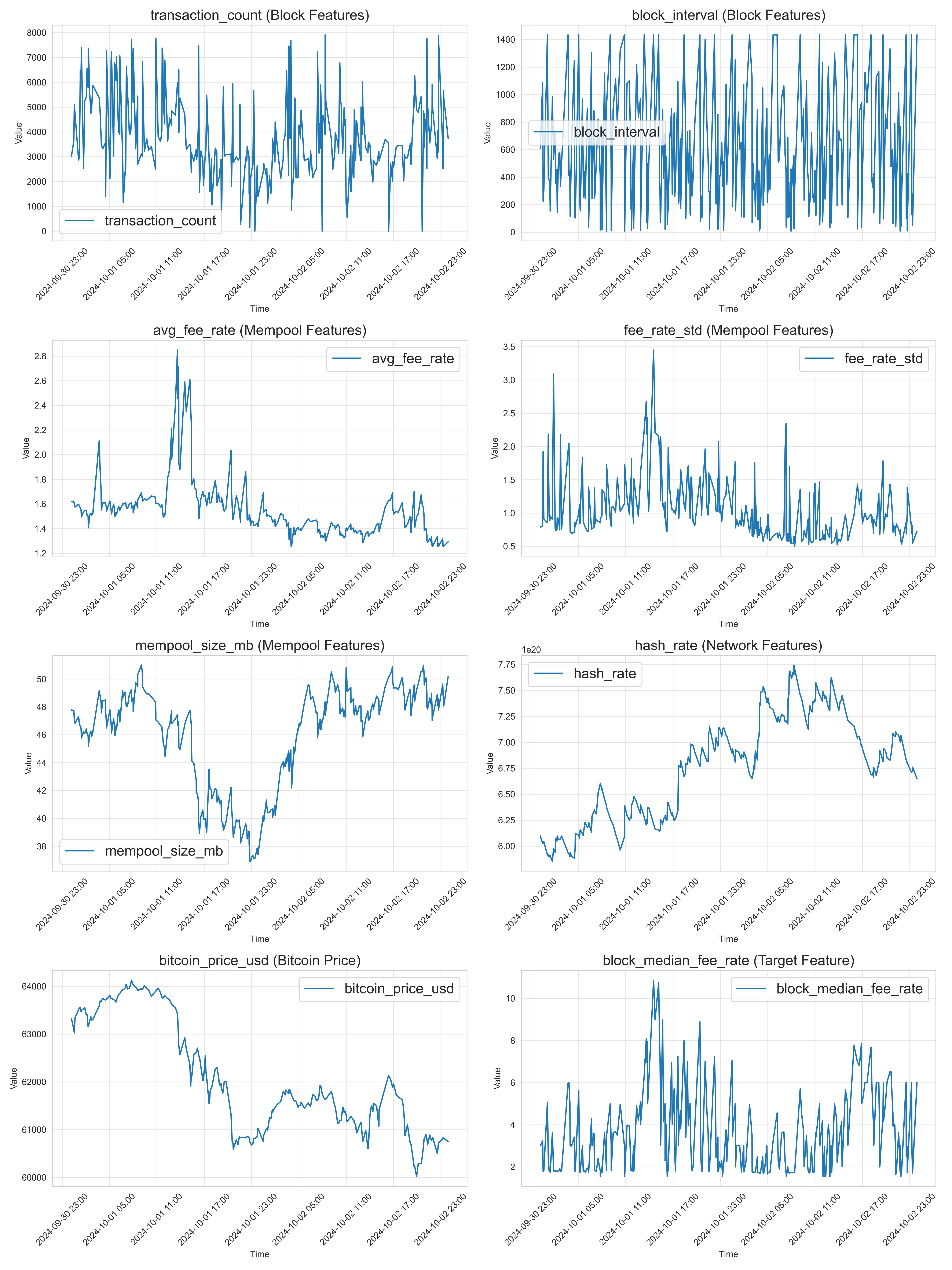}
\caption{Temporal dynamics of Bitcoin transaction activity, network state, and fee dynamics during a two-day observation period (2024-10-01 to 2024-10-02). The figure highlights significant variability in transaction counts, block intervals, mempool metrics, hash rate, and Bitcoin price, as well as the volatility of the block median fee rate. These patterns underscore the dynamic nature of the Bitcoin network and the interconnected relationships between congestion, fee-setting behaviors, and network security.}

\label{fig:bitcoin_metrics_revised}
\end{figure*}

\begin{figure*}[!t]
    \centering
    \includegraphics[width=\textwidth]{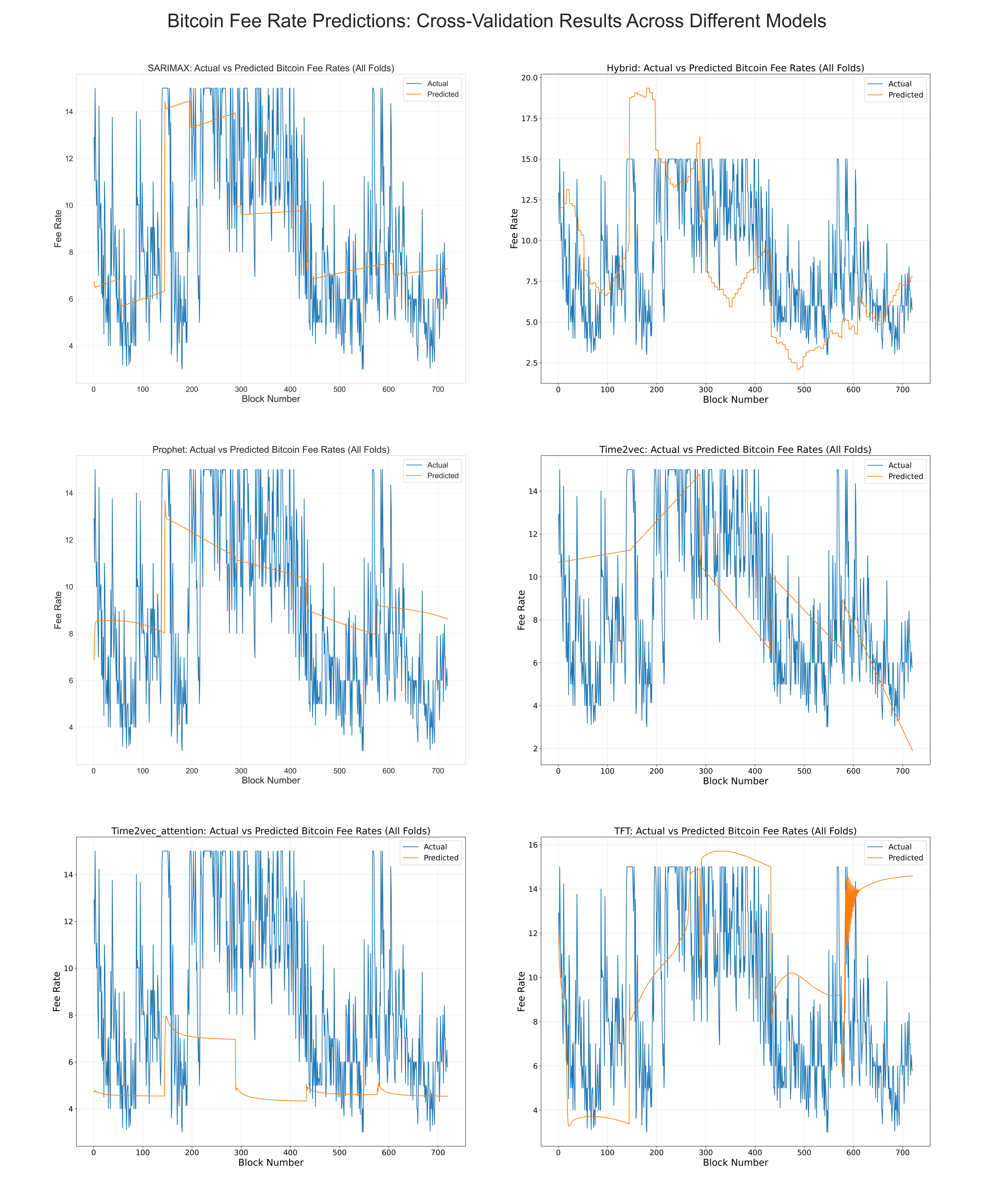}
    \caption{Comparison of Cross-Validation Results Across Multiple Time Series Models: SARIMAX, Prophet, Hybrid, Temporal Fusion Transformer (TFT), Time2Vec, and Time2Vec + Attention. Blue lines represent actual values, while orange lines show predicted fee rates across different validation folds.}
    \label{fig:model_comparison_appendix}
\end{figure*}

\begin{figure*}[!t]
    \centering
    \includegraphics[width=\textwidth]{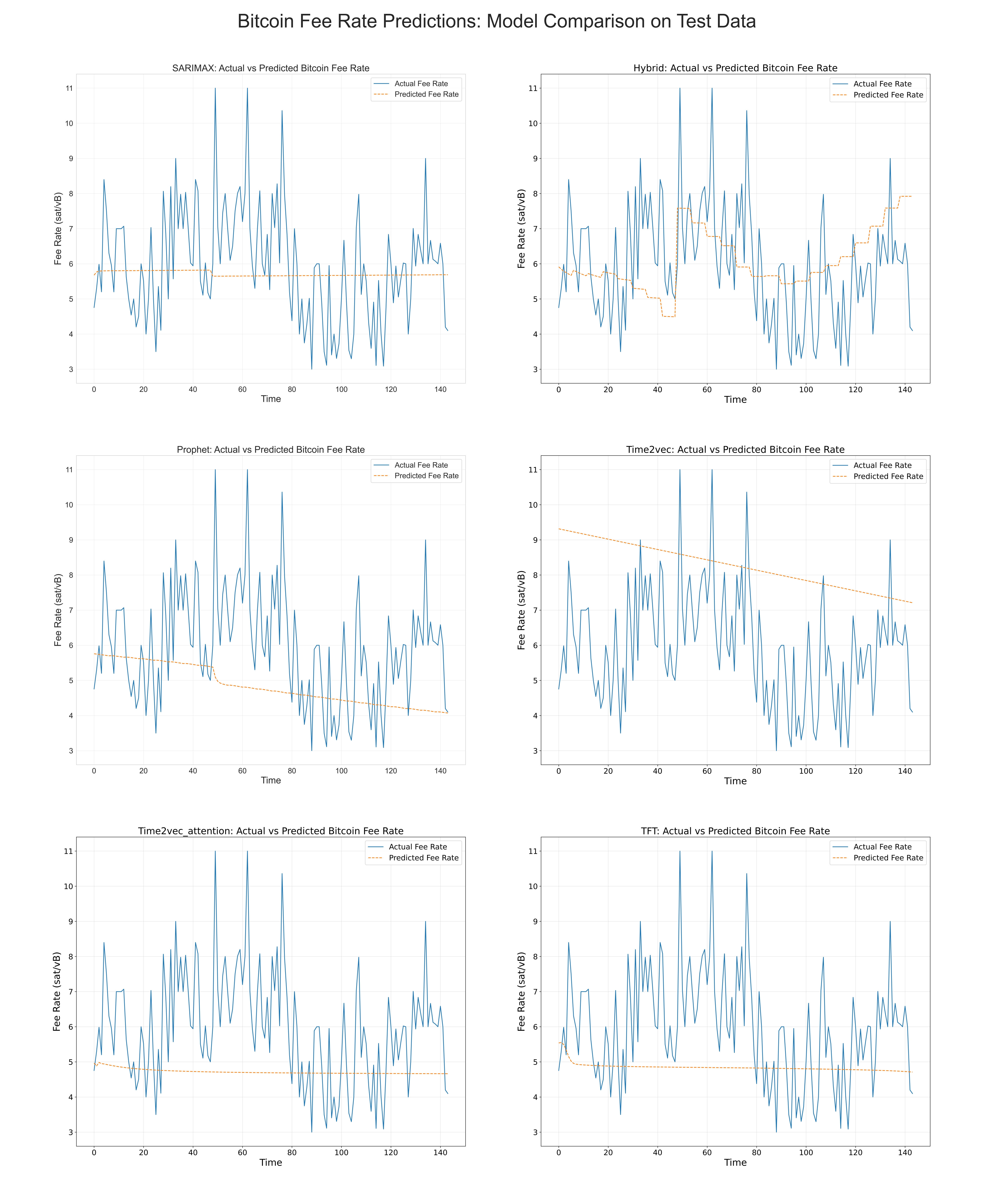}
    \caption{Comparison of Test Results Across Multiple Time Series Models: SARIMAX, Prophet, Hybrid, Temporal Fusion Transformer (TFT), Time2Vec, and Time2Vec + Attention. Results show model predictions over the 144-block test period, with blue lines indicating actual fee rates and orange lines showing predictions.}
    \label{fig:model_comparison_test_appendix}
\end{figure*}

\end{document}